\documentclass[runningheads]{llncs}
\usepackage[T1]{fontenc}
\usepackage{graphicx,verbatim}
\usepackage{booktabs,multicol,multirow}
\usepackage{amsmath, amsfonts, amssymb}
\usepackage{graphicx}
\usepackage{xcolor}

\usepackage[misc,geometry]{ifsym}

\newcommand{\BLUE}[1]{\textcolor{blue}{#1}}
\newcommand{\std}[1]{{\scriptsize$\pm$#1}}

\begin{document}

\title{Compass: Prostate Cancer Detection Needs Multi-View Context}

\author{Paul F.R. Wilson\inst{1,6}${^{\star}}$ \and
Mohamed Harmanani\inst{1,6}${^{\star}}$ \and
Zhuoxin Guo\inst{1,6} \and
Obed~K.~Dzikunu\inst{2,6} \and
Hannes Cash\inst{3} \and
Adam Kinnaird\inst{4} \and
Brian Wodlinger\inst{5} \and
Purang~Abolmaesumi\inst{2}$^\dagger$ \and
Parvin~Mousavi\inst{1,6}$^\dagger$}

\authorrunning{Wilson/Harmanani et al.}
\titlerunning{Compass: PCa Detection Needs Multi-View Context}

\institute{Queen's University, Kingston, Canada \and
University of British Columbia, Vancouver, Canada \and
Otto von Guericke University Magdeburg, Germany \and
University of Alberta, Edmonton, Canada \and
Exact Imaging, Markham, Canada \and
Vector Institute, Toronto, Canada}

\renewcommand{\thefootnote}{\fnsymbol{footnote}}
\footnotetext[1]{co-first authors (interchangeable order)}
\footnotetext[4]{co-senior authors}
\renewcommand{\thefootnote}{\arabic{footnote}}
\setcounter{footnote}{0}

\titlerunning{Compass: PCa Detection Needs Multi-View Context}
    
\maketitle              

\begin{abstract}
Artificial intelligence (AI) analysis of micro-ultrasound ($\mu$US) has shown promise for prostate cancer (PCa) detection. However, most existing AI methods focus on the analysis of single $\mu$US images in isolation. By contrast, expert $\mu$US readers typically assess a full recorded video study, which provides three-dimensional context, to improve PCa detection compared to single-frame analysis. Inspired by this clinical workflow, we propose Compass, a novel AI methodology which models a $\mu$US study as a stream of 2D images. Compass jointly integrates rotational sweep videos of the prostate with $\mu$US frames acquired at the moment of biopsy, and performs evidence aggregation across the study using a transformer conditioned on the probe’s rotational angle. Finally, a decoder head predicts frame-level and study-level risk scores for the patient. The model is trained and evaluated using a multi-center clinical trial dataset of $\mu$US studies, including continuous rotational scans of the prostate and videos captured during biopsy acquisition. We compare the proposed method to baseline AI methods from the literature and to risk scores provided by clinical experts. Our framework shows strong performance, highlighting the value of multi-view context for $\mu$US PCa detection, and providing a potentially powerful tool to complement human expertise in $\mu$US-based PCa diagnosis. Our code is available at: \BLUE{\url{https://github.com/mharmanani/Compass}}.


\end{abstract}

\section{Introduction}
Prostate cancer (PCa) is the second most commonly diagnosed cancer in men worldwide and one of the leading causes of cancer-related deaths~\cite{schafer2025recent}. Accurate early detection and risk stratification of PCa are essential to guide treatment planning and ensuring favorable outcomes. Medical imaging plays a crucial role in the clinical workflow by localizing suspicious regions to guide biopsies, after which histopathology provides the definitive diagnosis and grade that drive treatment decisions.\\ 
\indent High-resolution micro-ultrasound ($\mu$US) is a new and promising modality for PCa detection, offering real-time visualization and lesion localization at substantially higher spatial resolution than conventional ultrasound, with a lower cost than {multi-parametric MRI} (mpMRI). The PRI-MUS~\cite{ghai2016primus} protocol and scoring system assigns risk scores from $\mu$US by interpreting prostate echotexture, and has shown non-inferior performance to mpMRI PI-RADS~\cite{kinnaird2025optimum}. Despite these advances, $\mu$US interpretation of PRI-MUS score is subject to operator expertise and can suffer from significant inter-observer variability. To facilitate reliable lesion targeting across diverse clinical settings and workflows, there is a need for an objective, user-agnostic tool for $\mu$US analysis and PCa detection.\\
\indent Modern artificial intelligence (AI) methods have demonstrated strong diagnostic accuracy across medical imaging, fueled in part by the recent rise of foundation models (FMs). In PCa detection, FM-based approaches have led progress in mpMRI~\cite{shao2025mriptpca,lee2025provicnet}, as well as conventional US~\cite{harmanani2025cinepro}, temporal-enhanced US~\cite{elghareb2025proteus}, and $\mu$US~\cite{wilson2024prostnfound,wilson2026prostnfoundplus,imran2026ai}. Most FM-based methods in this space operate primarily on single-plane 2D B-mode images to produce e.g., per-image heatmaps and risk scores~\cite{wilson2026prostnfoundplus}. However, $\mu$US interpretation is inherently {three-dimensional} (3D) and \emph{spatially continuous}: clinicians collect cine-loops of systematic prostate sweeps to localize suspicious patterns across multiple views, and determine biopsy targets. While current literature in US-based PCa detection have explored the integration of information from cine-loops~\cite{harmanani2025cinepro}, temporal dynamics~\cite{elghareb2025proteus}, and heuristic slice-level aggregation into AI models~\cite{imran2026ai}, the analysis of full prostate sweeps remains largely untapped. There is a clear gap between how clinicians collect and interpret ultrasound data versus the state-of-the-art in AI-based PCa detection. \\
\indent To bridge this gap, we propose \textbf{Compass}, a multi-view framework that leverages the spatio-temporal structure of complete prostate studies for accurate cancer detection. Compass aggregates evidence from two sources: (i) cine-loops of $\mu$US sweeps of the prostate, recorded while the probe rotates around its main axis, providing rich multi-view context, and (ii) $\mu$US 2D biopsy frames. We evaluate our framework on a multi-center clinical trial dataset, comparing against prior baselines and expert risk assessments. Our main contributions are: \\
\indent 1. A multi-view reframing of $\mu$US PCa detection that leverages the rich clinical context from 3D cine-loops, enabling robust patient-level decision-making from spatially continuous evidence rather than isolated frames. \\
\indent 2. A novel architecture for mixed-evidence reasoning that jointly integrates 3D sweeps with 2D biopsy frames, conditioned on the ultrasound probe's rotational angle, effectively aligning \emph{global} context with \emph{local} evidence. \\
\indent 3. Validation on a new multi-center dataset to compare with human experts and state-of-the-art (SOTA) AI methods, establishing the practical value of AI-based multi-view decision support in $\mu$US PCa detection. 

\section{Materials}

\textbf{Data.}
We used a subset of data from the OPTIMUM~\cite{kinnaird2025optimum} clinical trial, comprising proprietary $\mu$US-guided prostate biopsy studies from two clinical centers: University of Alberta, Edmonton, Canada (UA) and Professionelle Urologie, Berlin, Germany (PU). The cohort includes 118 patients (76 UA, 42 PU). The ExactVu micro-ultrasound platform (Exact Imaging, Markham, Canada), which uses a high-frequency 29\, MHz side-firing linear-array transrectal probe (512 elements, 90\,$\mu$m pitch), was used. During each study, at least one initial transverse scan was performed by rotating the side-firing probe about its longitudinal axis, sweeping the imaging plane from the left lateral aspect of the prostate through midline to the right lateral aspect. Subsequently, a median of 15 biopsy samples per patient were acquired from systematic sextant biopsy locations and from any suspicious lesions identified by the performing urologist using the PRI-MUS~\cite{ghai2016primus} protocol. UA used transrectal while PU used transperineal biopsies. For each biopsy core, multi-frame B-mode cine sequences were recorded. For each core, the needle-trace region was annotated, and we use the frame immediately preceding needle firing as the canonical biopsy image for core-level modeling. Throughout the procedure, an inertial measurement unit continuously records the roll angle about the probe's longitudinal axis, providing an approximation of the left-right position of the imaging plane.

\textbf{Labels and clinical biomarkers.}
Histopathology provides per-core diagnosis and ISUP Grade Group (GG). For a biopsy sample we define clinically significant prostate cancer (csPCa) as GG$\geq 2$ for a biopsy core. For a patient we define csPCa as having at least one biopsy sample containing csPCa, following~\cite{kinnaird2025optimum}. The dataset also provides routinely available clinical biomarkers such as the patient's age and prostate specific antigen (PSA).

\section{Methods}
\begin{figure*}[t!]
    \centering
    \includegraphics[width=\textwidth]{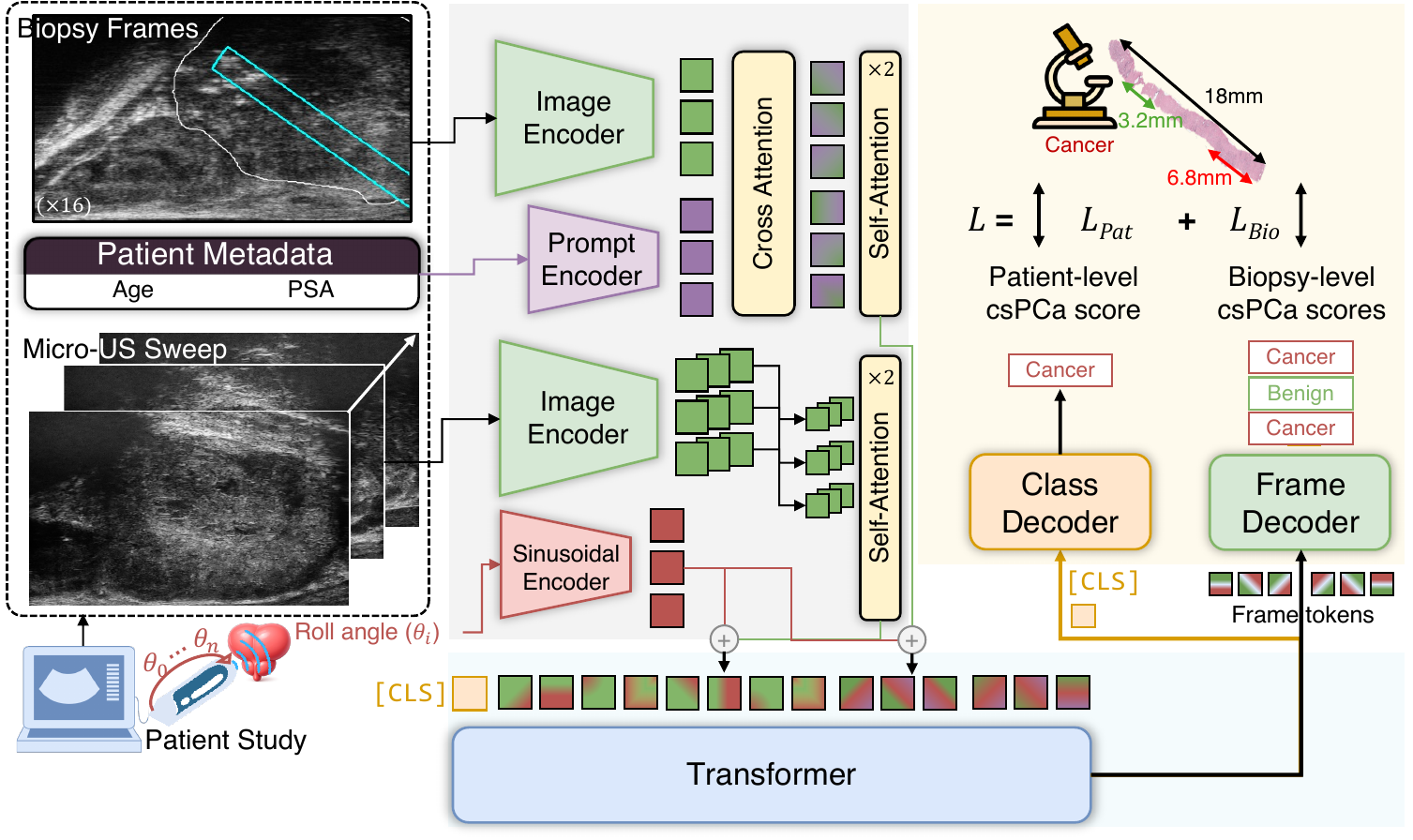}
    \caption{{\bf Compass:} Dual-branch patient-level framework that integrates two complementary sources of evidence: (i) rotational $\mu$US sweeps that provide dense multi-view coverage of the gland, and (ii) biopsy core frames paired with clinical metadata.}
    \label{fig:compass}
\end{figure*}

An overview of our method is shown in Figure~\ref{fig:compass}.\\
\indent\textbf{Sweep-level feature extraction.}
Each $\mu$US rotational sweep is treated as a set of frames with known transducer roll angles. 
Let $\{(F_t,\theta_t)\}_{t=1}^{T}$ denote the selected sweep frames and their roll angles.
A frozen image encoder produces per-frame visual tokens $U_t = \Phi(F_t) \in \mathbb{R}^{M\times d}$,
where $M$ is the number of image tokens and $d=256$. We adopt ProstNFound$^+$~\cite{wilson2026prostnfoundplus} as the backbone, due to its strong performance on and robust prostate US representations. 

To encode acquisition geometry, we embed the continuous roll angle with a sinusoidal encoder $r_t = \mathrm{PE}(\theta_t) \in \mathbb{R}^{d}.$ We then summarize each frame using a lightweight attention module $\mathrm{Dec}_{\text{sw}}(\cdot)$, and add the roll embedding and shared sweep token embedding $\tau^\mathrm{sw}$ to obtain a sweep token $z^{\mathrm{sw}}_t$ per timestep:
\[
z^{\mathrm{sw}}_t = \mathrm{Dec}_{\mathrm{sw}}\big(U_t\big) + r_t + \tau^{\mathrm{sw}}\in \mathbb{R}^{d}.
\]

\textbf{Biopsy-level frame encoding.}
We denote the biopsy frames and their roll angles by $\{(I_i,\theta_i)\}_{i=1}^{N}$. Following ProstNFound$^+$, we embed the biomarker vector $m=(\text{Age},\text{PSA})$ using a prompt encoder $\Pi$ that maps normalized scalars to a small set of $K$ learnable prompt tokens $P=\Pi(m)\in\mathbb{R}^{K\times d}$.

We encode each biopsy frame and roll angle with the same frozen ProstNFound$^+$ encoder and positional embedding module to obtain visual tokens and roll embeddings $V_i=\Phi(I_i)\in\mathbb{R}^{M\times d}$ and $r_i = \mathrm{PE}(\theta_i)$.\\
\indent We then fuse clinical prompts with image evidence using a lightweight decoder $\mathrm{Dec}_{\mathrm{bx}}(\cdot)$ that performs self-attention and cross-attention between the visual tokens and the prompt set. The roll angle embedding and shared biopsy token embedding $\tau_\mathrm{bx}$ is added, resulting in the per-biopsy tokens $z^\mathrm{bx}_i \in \mathbb{R}^d$:
\[
z^{\mathrm{bx}}_i = \mathrm{Dec}_{\mathrm{bx}}\big(V_i,\;P\big) + r_i +\tau_{\mathrm{bx}} \in \mathbb{R}^{d}.
\]
\textbf{Mixed-token transformer reasoning.}
Given biopsy tokens $\{z^{\mathrm{bx}}_i\}_{i=1}^{N}$ and sweep tokens $\{z^{\mathrm{sw}}_t\}_{t=1}^{T}$, we form a single mixed-token sequence by concatenation and prepend a learnable \texttt{[CLS]} token:
\[
X = [\texttt{CLS},\, z^{\mathrm{bx}}_1,\dots,z^{\mathrm{bx}}_N,\, z^{\mathrm{sw}}_1,\dots,z^{\mathrm{sw}}_T] \in \mathbb{R}^{(1+N+T)\times d}.
\]
We pass the sequence through a transformer encoder $\mathrm{Trans}(\cdot)$ that applies global multi-head self-attention over \emph{all} tokens:
\[
H = \mathrm{Trans}(X), \qquad H \in \mathbb{R}^{(1+N+T)\times d}.
\]
This design enables cross-branch evidence exchange: biopsy tokens can contextualize sparse biopsy cues with broader gland coverage, while sweep tokens can emphasize regions that are most consistent with biopsy-confirmed malignancy. The patient-level score is predicted from the \texttt{[CLS]} state using a class decoder head, and biopsy-level scores are predicted from the biopsy token states using a frame/biopsy decoder head. 
\[
\hat{y}_{\mathrm{pat}} = h_{\mathrm{pat}}(H_{\texttt{CLS}}), 
\qquad
\hat{y}^{(i)}_{\mathrm{bio}} = h_{\mathrm{bio}}(H_i),\; i=1,\dots,N.
\]

\textbf{Loss function.}
We train Compass with a joint objective combining patient-level supervision and biopsy-level auxiliary supervision: 
\[
\mathcal{L}
= \mathcal{L}_{\mathrm{Pat}} + \lambda\,\mathcal{L}_{\mathrm{Bio}}
= \mathrm{CE}\!\left(\hat{\mathbf{y}}_{\mathrm{pat}}, y_{\mathrm{pat}}\right)
+ \frac{\lambda}{N}\sum_{i=1}^{N}\mathrm{CE}\!\left(\hat{\mathbf{y}}^{(i)}_{\mathrm{bio}}, y^{(i)}_{\mathrm{bio}}\right),
\]
where $\lambda$ is a constant weighting hyperparameter.

\section{Experiments}

\textbf{Baselines \& Evaluation.}
We compare Compass to three baseline families: (i) frame-level classifiers (CLIP ViT-L/14, MedSAM, Cinepro, ProstNFound$^+$~\cite{wilson2026prostnfoundplus}); (ii) {multi-instance learning} (MIL) variants using transformer aggregation over frame/core embeddings (Cinepro-MIL, MedSAM-MIL, ProstNFound$^+$-MIL); and (iii) video models, including ViViT~\cite{arnab2021vivit}, 3D-ResNet18, and hybrid encoder– transformer models (e.g., MicroSegnet~\cite{jiang2024microsegnet}, ProstNFound$^+$ + ViViT). All experiments used 5-fold patient-level cross-validation. We report patient-level AUROC, balanced accuracy, and sensitivity at 60\% specificity (Sen@60), as well as biopsy-level AUROC and balanced accuracy at the epoch with the best mean patient AUROC across folds.

\indent\textbf{Implementation Details.} All experiments were conducted in PyTorch using a single NVIDIA GPU with mixed-precision training. We optimized all models with AdamW (weight decay $10^{-4}$). Compass was trained for 100 epochs with a constant learning rate of $3\times10^{-5}$ and a 5-epoch linear warmup, using a batch size of 2. The backbone was frozen during training, gradient clipping was set to 1.0, and the token-level loss weight $\lambda$ was 0.05.

\section{Results and Discussion}

\begin{table*}[t!]
\caption{Patient and biopsy-level classification performance across relevant baselines. We report the mean and standard deviation over 5-fold cross-validation.}
\centering
\setlength{\tabcolsep}{5pt}
\begin{tabular}{l|ccc|cc}
\toprule
\multirow{2}{*}{\bf Method}
  & \multicolumn{3}{c|}{\bf Patient-level}
  & \multicolumn{2}{c}{\bf Biopsy-level} \\
\cmidrule{2-4}\cmidrule{5-6}
  & \bf AUROC & \bf B.Acc. & \bf Sen@60
  & \bf AUROC & \bf B.Acc. \\
\midrule
CLIP ViT-L/14
  & $67.9$\std{7.9}  & $71.3$\std{5.3}  & $57.9$\std{18.0}
  & $62.6$\std{11.4} & $62.9$\std{8.0} \\
Cinepro
  & $72.6$\std{6.5}  & $76.9$\std{8.3}  & $68.2$\std{13.1}
  & $68.6$\std{7.7}  & $67.6$\std{4.2} \\
MedSAM
  & $74.6$\std{5.8}  & $76.8$\std{5.3}  & $43.1$\std{24.4}
  & $61.7$\std{4.3}  & $61.6$\std{3.1} \\
ProstNFound$^+$
  & $78.7$\std{15.4} & $82.9$\std{12.5} & $76.7$\std{17.8}
  & $68.8$\std{10.3} & $65.9$\std{6.6} \\
\midrule
Cinepro-MIL
  & $76.3$\std{13.9} & $76.1$\std{11.2} & $73.3$\std{19.2}
  & -- & -- \\
MedSAM-MIL
  & $75.7$\std{13.7} & $76.4$\std{10.3} & $73.3$\std{19.2}
  & -- & -- \\
ProstNFound$^+$-MIL
  & $77.3$\std{14.5} & $78.1$\std{9.4} & $82.5$\std{15.3}
  & -- & -- \\
\midrule
ViViT
  & $61.9$\std{13.9} & $66.0$\std{10.4} & $61.7$\std{23.1}
  & -- & -- \\
3D-ResNet18
  & $71.1$\std{6.0}  & $72.3$\std{4.6}  & $73.2$\std{11.9}
  & -- & -- \\
MicroSegNet+ViViT
  & $67.7$\std{7.9}  & $69.0$\std{5.2}  & $66.8$\std{11.8}
  & -- & -- \\
ProstNFound$^+$+ViViT
  & $66.4$\std{15.3} & $68.8$\std{9.5}  & $66.6$\std{17.6}
  & -- & -- \\
\midrule
PRI-MUS {(Experts)}
  & $78.5$\std{11.1} & $78.3$\std{6.9} & $83.8$\std{15.8}
  & ${\bf 73.1}$\std{8.9} & ${\bf 71.9}$\std{7.5} \\
\midrule
{\bf Compass} (ours)
  & ${\bf 87.2}$\std{8.7} & ${\bf 85.0}$\std{8.2} & ${\bf 89.9}$\std{10.0}
  & ${\underline{70.5}}$\std{8.6} & ${\underline{68.3}}$\std{6.9} \\
\bottomrule
\end{tabular}
\label{tab:results1}
\end{table*}

\textbf{Diagnostic Performance.} Table~\ref{tab:results1} compares our performance against several prominent baselines from the literature. 
Among baselines that use 2D biopsy planes only, CLIP is the weakest, likely because it has been trained mainly with non-ultrasound data. Cinepro exhibits moderate performance at the core-level due to its robust training strategy, but falls behind MedSAM at the patient-level. Interestingly, MedSAM obtains very low sensitivity, highlighting the weaknesses of naive FM fine-tuning and the importance of domain-specific training strategies. Finally, ProstNFound$^+$ achieves the highest performance among baselines ($+4.1\%$ Patient AUROC, $+12.9\%$ Sen@60) since it integrates relevant domain knowledge via prompts~\cite{wilson2026prostnfoundplus,wilson2024prostnfound}.\\
\indent Compared to frame-level baselines, MIL primarily improves sensitivity by aggregating evidence over multiple instances, reducing per-frame noise and stabilizing threshold behavior. For example, MedSAM-MIL yields a large gain in sensitivity (+$30\%$) over MedSAM, and ProstNFound$^+$-MIL substantially improves sensitivity (+$5.8\%$) over its single-frame variant. This highlights the benefit of set-based aggregation in biopsy workflows: by pooling evidence across multiple frames, MIL reduces noise and sampling variance, and better reflects the clinical reality that csPCa cues are sparse and heterogeneous across the prostate gland. The video baselines underperform compared to MIL ($\approx7\%$ Patient AUROC, Bal.Acc. for ProstNFound$^+$+ViViT), which is expected in low-data regimes, as spatio-temporal models are data hungry and struggle under weak supervision.\\
\indent Finally, Compass achieves the {highest \emph{patient-level} performance compared to other baselines in this cohort}, due to several key reasons: (i) it leverages a strong clinical backbone to produce stable per-image evidence tokens, and (ii) incorporates sweep information rather than treating frames independently. Crucially, (iii) Compass performs \emph{cross-branch reasoning} by fusing biopsy and sweep tokens with roll-aware conditioning in a single transformer. This lets Compass align localized biopsy evidence with rotation-consistent sweep patterns, yielding a large improvement over ProstNFound$^+$, MIL baselines, and generic video models. {With respect to \textit{core-level} performance, Compass exhibits comparable but lower performance than PRI-MUS 
(-2.6\% AUROC). This suggests that 
Compass's value lies not in replacing PRI-MUS, but 
in complementing it through comprehensive multi-view context that core-level methods do not capture}.\\
\begin{table}[t]
\footnotesize
\caption{Ablation experiments at the patient-level show each component's importance.}
\centering
\begin{tabular*}{\textwidth}{@{\extracolsep\fill}lccccc@{}}
\toprule
& \multicolumn{3}{c}{\bf Components} & & \\
\cmidrule{2-4}
\bf Method & \bf Sweeps & \bf Angle PE & \bf Transformer & {\bf AUROC} & {\bf Bal.Acc.} \\
\midrule
No angle embedding  & $\checkmark$ & $\times$ & $\checkmark$ & 82.7 & 75.1 \\
No transformer     & $\checkmark$ & $\checkmark$ & $\times$ & 74.3 & 66.2 \\
No sweeps       & $\times$ & $\checkmark$ & $\checkmark$ & 78.8 & 70.6 \\
\midrule
{\bf Compass} (ours)                               & $\checkmark$ & $\checkmark$ & $\checkmark$ & {\bf {87.2}} & {\bf {85.0}} \\
\bottomrule
\end{tabular*}
\label{tab:ablation}
\end{table}
\indent\textbf{Ablation Study.} Table~\ref{tab:ablation} isolates the contribution of key components in the Compass framework. Removing the roll-angle positional encoding reduces performance (-4.8\% AUROC, -1.4\% Bal.Acc.), indicating that encoding probe rotation allows the model to exploit sweep structure beyond treating frames as an unordered set. Removing the transformer causes the largest degradation (-13.2\% AUROC, -10.3\% Bal.Acc.), underscoring that \emph{cross-branch reasoning} is critical: integrating sweep and biopsy evidence across branches is key to Compass' success. Finally, the biopsy-only ProstNFound$^+$ variant underperforms Compass by 8.7\% in AUROC and 5.9\% in Bal.Acc., confirming that sweep information provides substantial complementary signal at the patient level. Overall, these ablations suggest Compass’s gains arise from the combination of (i) sweep evidence, (ii) rotation-aware structure, and (iii) transformer-based fusion that enables biopsies and sweep frames to condition one another.\\
\begin{figure*}[t!]
    \centering
    \includegraphics[width=0.9\textwidth]{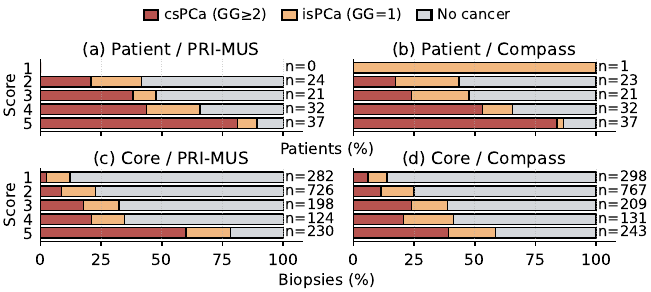}
    \caption{Outcome distributions across score bins for PRI-MUS and Compass at the patient (top) and core (bottom) levels. Bars show percentages of csPCa, insignificant PCa, and benign cases; $n$ indicates bin counts.}
    \label{fig:compass_vs_primus}
\end{figure*}
\indent\textbf{Clinical evaluation.} Figure~\ref{fig:compass_vs_primus} summarizes how csPCa prevalence varies across score strata for PRI-MUS and Compass at both the patient and core levels, complementing the scalar metrics in Table~\ref{tab:results1}. At the patient level, Compass produces a clear monotonic risk gradient: high-score bins concentrate a larger fraction of csPCa, while low-score bins are dominated by benign and clinically insignificant disease, indicating more effective study-level separation under the same clinical burden. This is reflected by Compass’s higher patient-level AUROC and improved Sen@60 score in Table~\ref{tab:results1}, suggesting it can prioritize more csPCa-positive patients while maintaining a similar false-positive burden. In contrast, PRI-MUS exhibits greater overlap between csPCa and non-csPCa categories across intermediate bins, consistent with weaker patient-level prioritization. \\
\indent At the biopsy core level, higher scores are associated with larger percentages of csPCa for both Compass and PRI-MUS. PRI-MUS exhibits a sharper concentration of csPCa at higher scores, and a clearer benign separation at lower scores, corresponding to its higher core-level AUC. This {reinforces Compass's complementary role in the clinical workflow}: Compass is best suited for sweep-informed patient triage, while PRI-MUS remains more informative for localized core-level prediction.

\begin{figure*}[t!]
    \centering
    \includegraphics[width=0.95\textwidth]{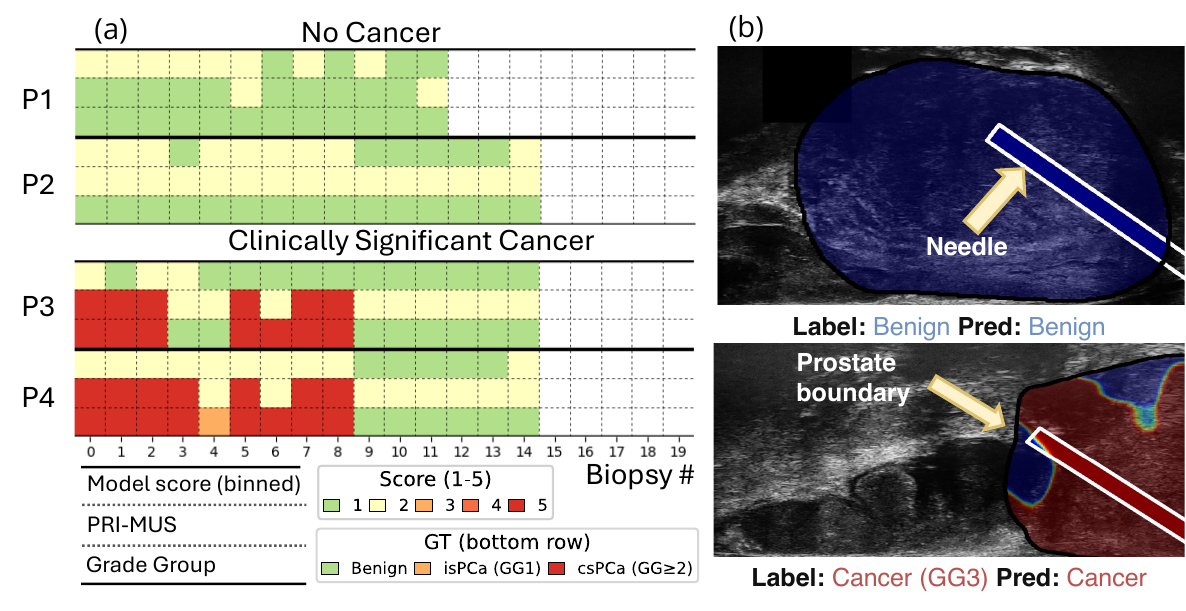}
    \caption{(a) Model and PRI-MUS score distribution compared to grade group ground truth across biopsies (left to right) for patients with no cancer (top 2 row groups) and csPCa (bottom two row groups). (b) Image-level cancer detection via post-hoc heatmaps conditioned on Compass embeddings.}
    \label{fig:heatmaps}
\end{figure*}
\indent\textbf{Qualitative evaluation.} To highlight the efficacy of Compass for patient-level analysis, we show the distribution of its risk scores across biopsies \emph{within} patients (Figure~\ref{fig:heatmaps}a). Compass's biopsy risk scores typically align with PRI-MUS risk scores and with ground truth pathology findings. We also generate post-hoc heatmaps by training an off-the-shelf mask decoder on top of our model's frame tokens. Figure~\ref{fig:heatmaps}b shows two representative biopsy cores, where the benign core (upper) has lower activations than the csPCa core (lower).

\section{Conclusion}
We introduced Compass, a study-level framework for $\mu$US prostate cancer detection that integrates two complementary evidence streams at biopsy time: rotational sweeps and biopsy cores. Compass substantially outperforms frame-level, MIL, and end-to-end video baselines. {By explicitly modeling sweep context, Compass shifts AI-based $\mu$US analysis from isolated-frame scoring toward study-level risk stratification, aligning more closely with how clinicians interpret full examinations. This opens the door to practical decision support for patient triage and targeted biopsy planning.\\

\noindent \textbf{Acknowledgments.} This work was supported by the Natural Sciences and Engineering Research Council of Canada (NSERC) and the Canadian Institutes of Health
Research (CIHR). Parvin Mousavi is supported by the Canada CIFAR AI Chair program, a Canada Research Chair, and the Vector
Institute.\\

\noindent \textbf{Disclosure of Interests.} Brian Wodlinger is VP Clinical and Engineering at Exact Imaging. Other authors have no potential conflicts of interest to disclose.

\bibliographystyle{splncs04}
\bibliography{refs}

\end{document}